\pdfoutput=1
\UseRawInputEncoding

\documentclass[a4paper, 10pt, conference]{ieeeconf}      

\IEEEoverridecommandlockouts                              

\usepackage{graphics} 
\usepackage{bm}
\usepackage{float} 
\usepackage{stfloats}
\usepackage{caption}
\usepackage{cancel}
\usepackage{widetext}
\usepackage{epsfig} 
\usepackage{overpic}
\usepackage{times} 
\usepackage{booktabs}
\usepackage{rotating}
\usepackage{algorithmic}
\usepackage{algorithm}

\usepackage{amsmath} 
\usepackage{amssymb}  

\usepackage{amsthm}
\usepackage{mathrsfs}
\usepackage[table, dvipsnames, svgnames, x11names]{xcolor} 
\usepackage{colortbl}
\usepackage{multirow}
\usepackage{marvosym}
\definecolor{mygray}{gray}{.9}
\usepackage{pifont}
\usepackage{cite}
\let\labelindent\relax
\usepackage{enumitem}
\usepackage[colorlinks,
            linkcolor=blue,
            anchorcolor=blue,
            citecolor=green,
            urlcolor=violet,
            backref=page]{hyperref}

\title{\LARGE \bf
Bridging the Resource Gap: Deploying Advanced Imitation Learning Models onto Affordable Embedded Platforms
}

\author{Haizhou Ge$^{1*}$, Ruixiang Wang$^{2*}$, Zhu-ang Xu$^{3*}$, Hongrui Zhu$^{4}$, Ruichen Deng$^{6}$, \\Yuhang Dong$^{4}$, Zeyu Pang$^{5}$,  Guyue Zhou$^{1}$, Junyu Zhang$^{5}$, Lu Shi$^{1\dagger}$
\thanks{*These authors contributed equally to this work. $^{\dagger}$Corresponding author: shilu@air.tsinghua.edu.cn, $^{1}$Tsinghua University, China, $^{2}$Harbin Institute of Technology, China, $^{3}$Jiangnan University, China, $^{4}$Zhejiang University, China, $^{5}$Tongji University, China, $^{6}$University of Pennsylvania, USA}\\
\thanks{We gratefully acknowledge the support of Discover Robotics and Horizon Robotics. Any opinions, findings, conclusions or recommendations expressed in this material are those of the authors and do not necessarily reflect the views of the funding agencies.}
}

\begin{document}

\maketitle
\thispagestyle{empty}
\pagestyle{empty}

\begin{abstract}
Advanced imitation learning with structures like the transformer is increasingly demonstrating its advantages in robotics. However, deploying these large-scale models on embedded platforms remains a major challenge. In this paper, we propose a pipeline that facilitates the migration of advanced imitation learning algorithms to edge devices. The process is achieved via an efficient model compression method and a practical asynchronous parallel method Temporal Ensemble with Dropped Actions (TEDA) that enhances the smoothness of operations. To show the efficiency of the proposed pipeline, large-scale imitation learning models are trained on a server and deployed on an edge device to complete various manipulation tasks.
\end{abstract}

\section{Introduction}
    

The recent advancements in transformer architectures~\cite{vaswani2017attention, Brohan2022RT1RT, Kim2021TransformerbasedDI} and large-scale models~\cite{ahn_as_2022, chi_diffusion_2023} have significantly propelled the field of embodied intelligence, enabling agents to interact with their environments in increasingly sophisticated ways. Among these advancements, imitation learning has emerged as a promising approach for addressing complex manipulation tasks~\cite{rakelly2019efficient,zare_survey_2023,radosavovic2021state}. By allowing agents to learn from expert demonstrations, imitation learning offers an efficient pathway to acquiring intricate behaviors without requiring extensive trial and error. The increasing reliance on imitation learning highlights its potential to overcome the challenges faced by traditional reinforcement learning methods, especially in environments that demand real-time decision-making and fine motor control~\cite{arulkumaran2017deep}. This has led to its growing adoption in robotic manipulation~\cite{Zhao2023LearningFB}, where rapid adaptability and high precision are critical. 

However, the deployment of these advanced algorithms often requires expensive computing platforms with rich resources~\cite{ke2021grasping, Duan2017OneShotIL}, which poses significant challenges when attempting to implement them in resource-limited environments. Examples of such environments include autonomous drones~\cite{azar2021drone}, mobile robots~\cite{fu_mobile_2024}, and other low-power, embedded systems where computational resources are constrained, and energy efficiency is paramount. The reliance on powerful hardware limits the accessibility and practicality of these large-scale models in many real-world applications.

Existing solutions to mitigate these challenges, such as cloud-based inference, have their limitations. While offloading computation to remote servers can reduce the local processing burden, it introduces dependencies on network connectivity. These dependencies can lead to latency issues, potential privacy concerns, and reliability problems in unstable network conditions~\cite{guo2018cloud}. The drawbacks highlight the need for alternative approaches that can operate effectively within the constraints of resource-limited platforms, leading to the growing interest in Edge AI, which refers to the deployment of artificial intelligence algorithms and models directly on edge devices. However, their limited computational power, storage, and memory make it challenging to deploy sophisticated algorithms or models directly onto edge devices.

\begin{figure*}[t]
    \centering
    \includegraphics[width=0.9\textwidth]{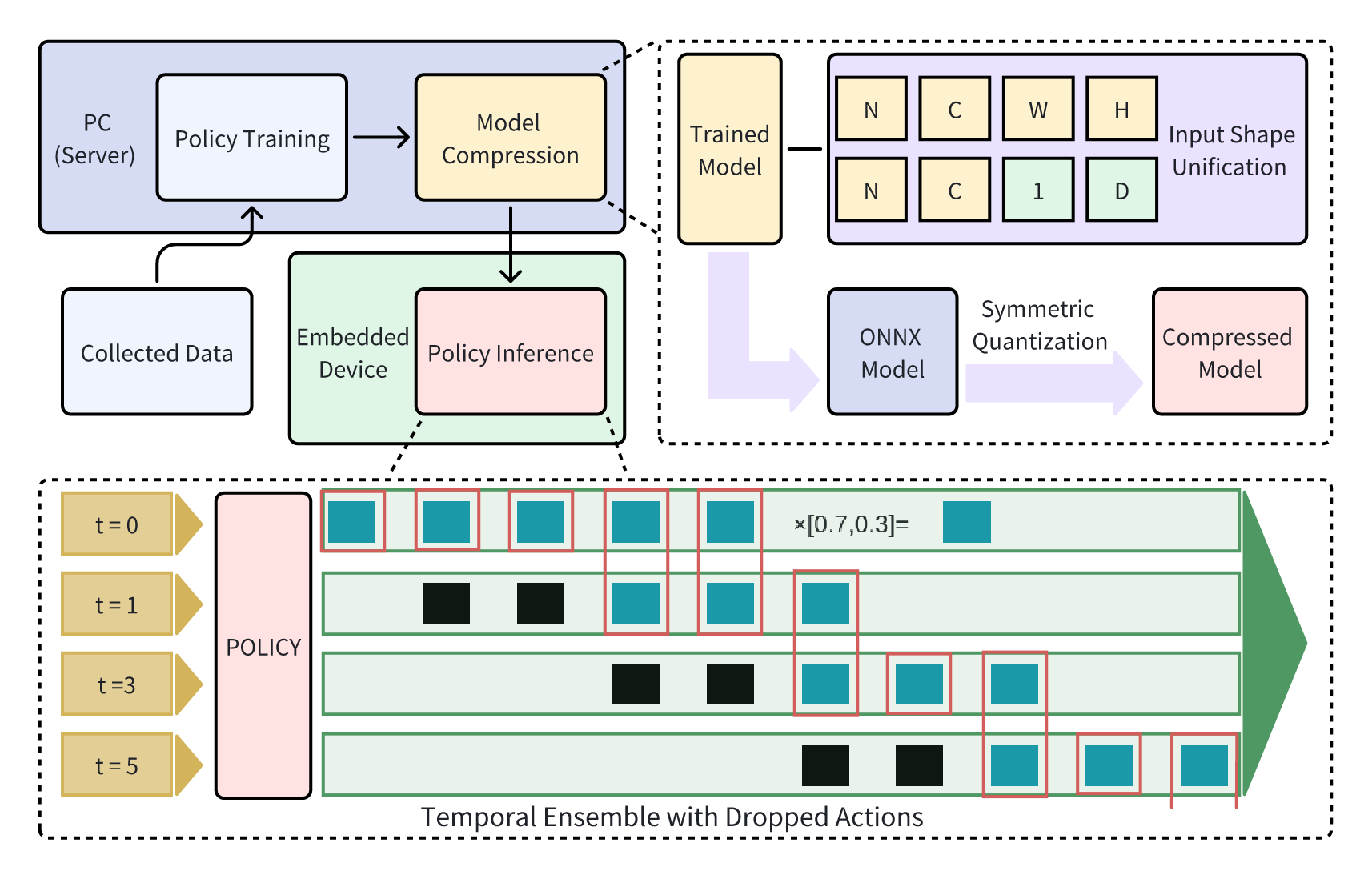}
    \caption{Pipeline for deploying Imitation Learning algorithms on embedded devices. The policy trained with collected data is deployed into the embedded device through two key components: model compression and Temporal Ensemble with Dropped Actions (TEDA)  with a trunking size of 5 for policy inference ($k=5$). In the TEDA section, the blue block represents the normal action, while the black block represents the dropped action and the red box represents Temporal Ensemble when applying actions.}
    \label{fig:1}
\end{figure*}

In this paper, we propose a pipeline (as shown in Fig.~\ref{fig:1}) to implement advanced imitation learning algorithms and models on low-cost embedded platforms to address the aforementioned problem. To bridge the cost gap and achieve a smoother transfer of the learning model from high-performance platforms to edge devices, we provide techniques for efficient model compression through input shape unification as well as parameter quantization. In the meantime, a practical asynchronous parallel method Temporal Ensemble with Dropped Actions (TEDA) is introduced to achieve high efficiency and avoid discontinuity during deployment. To validate the approach, we developed a complete system and conducted a series of robotic manipulation tasks involving both single-arm and dual-arm operations. 



\section{Related Work}
\subsection{Embodied Intelligence System}
Embodied intelligence systems aim to develop cognitive abilities and further motor skills through interactions with the environment~\cite{duan2022survey}. Both the models for single modality such as visual~\cite{Zeng2020TransporterNR} as well as language representations~\cite{Devlin2019BERTPO} and multimodal models have been introduced for robotic implementations, improving task decomposition, inference, and control~\cite{huang2023voxposer,shridhar2022cliport,stone2023open}, enabling more advanced robotic behavior. However, deploying these models for complex tasks poses significant challenges, including model deployment, accurate sensor data acquisition, and precise calibration, all of which drive up system costs. To address these challenges, cost-effective solutions like  ALOHA~\cite{zhao_learning_2023} have been developed, enabling end-to-end imitation learning directly from real demonstrations gathered using a custom teleoperation interface. These solutions highlight the ongoing efforts to balance the sophistication of AI models with the practical constraints of deployment in real-world environments.

\subsection{Edge AI}
Edge AI refers to the deployment of AI models and enables them to run directly on edge devices, close to the data source\cite{liu2022bringing}. Edge AI significantly reduces latency, decreases bandwidth usage, and enhances reliability by minimizing dependence on cloud connectivity. It has been applied across various domains, including autonomous driving~\cite{katare2023survey}, smart surveillance~\cite{ke2020smart}, and other IoT devices~\cite{merenda2020edge}. However, optimizing AI models for low-power devices and effectively utilizing limited computational resources remain primary challenges~\cite{singh2023edge}. Recent advancements in model compression techniques, such as pruning and quantization, have made it possible to deploy sophisticated AI algorithms on edge devices~\cite{reidy2023work, chen2020gpu}, which provides the way for achieving high inference accuracy in resources-limited devices. 

\section{Method}
In this section, we introduce the comprehensive pipeline (as shown in Fig.~\ref{fig:1}) to deploy advanced imitation learning algorithms into affordable embedded platforms. Firstly, policies are trained using the collected data for each task. Then, after unifying the input shapes, the model undergoes a compression phase via Symmetric Quantization (SQ). Following compression, the optimized model is deployed onto the embedded device, where it performs inference efficiently by utilizing TEDA. This technique not only improves the model's responsiveness but also ensures that it can make real-time decisions in dynamic environments. The subsequent subsections provide a detailed explanation of each stage in the pipeline, highlighting the methods and techniques employed to achieve robust performance on embedded platforms.


\subsection{Model Compression}
Due to the significant disparity in storage capacity and computational power between server-grade hardware and edge devices, directly deploying advanced models trained on servers to edge devices is impractical. To overcome this limitation, model compression is necessary. Before the compression process, we first apply an Input Shape Unification (ISU) technique. The inputs for the intelligent embodiments usually enjoy multi-modalities with different data structures, which leads to extra effort for storage and management. To deal with the problem, we express all the modalities into the typical image tensor form of N×C×H×W, where N represents the batch size, C represents the number of color channels, H and W describe the height and width of the image. For example, a joint position input with the shape of 1×D can be viewed as an image tensor with one color channel, i.e., its shape can be represented as 1×1×1×D. By using this consistent formation, redundant data transmission~\footnote{The inconsistent formation can lead to operators that are unsupported by the edge AI acceleration chip and can only run on the CPU. This would result in repeated data transfers between the AI chip and the CPU, ultimately causing a significant decrease in inference speed.} and extra computational overhead can be reduced. 

Then, we compressed the model by converting its parameters from 32-bit floating-point (float) to 16-bit integer (int16) using Symmetric Quantization (SQ). This approach reduces the model's memory footprint and computational requirements, enabling more efficient inference on resource-constrained embedded processors. By employing symmetric quantization, which uses a uniform scaling factor across all values, we maintain computational simplicity while significantly enhancing the model's deployability on low-power hardware without sacrificing too much accuracy.

\subsection{Temporal Ensemble with Dropped Actions (TEDA)}
Action chunking, which groups individual actions together and executes them as one unit, effectively reduces cumulative error in IL but can introduce motion jitter. Temporal Ensemble (TE), which performs a weighted average over predicted actions at the same timestep, can smooth out this jitter, yet it requires inference before each action. Those are two key techniques used in recent advanced imitation learning algorithms~\cite{zhao_learning_2023, Diffusion_Policy}. However, on embedded platforms, where the inference frequency is close to and even lower than the action execution frequency, inference before each action can cause discontinuous motion. This discontinuity not only increases the completion time but also tends to cause jitter, which reduces the success rate of the task.

To solve this problem, we propose a simple yet novel approach TEDA, which parallelizes the execution of action sequences and policy prediction. As shown in Fig.~\ref{fig:1} and Alg.~\ref{alg:teda_inference}, this approach performs one prediction at $t = 0$ $\left(t_{0}\right)$ to obtain $k$ predicted actions $\hat{a}_{0:k}^{t_0}$, then executes the first action $\hat{a}_{0}^{t_0}$ and performs another prediction at $t_{1}$. During the prediction process, the subsequent actions $\hat{a}_{1:N}^{t_0}$ ($N <= k$) are executed in parallel. Then at $t_{N+1}$, the prediction is finished and we get the newly predicted actions $\hat{a}_{1:k+1}^{t_1}$. Since the subsequent actions $\hat{a}_{1:N}^{t_1}$ lag behind the current time step $t_{N+1}$, they will not be executed, which we called dropped actions. The prediction and execution of actions will continue into the next cycle.

Essentially, a dropped action is an action that has not been predicted at a specific time step. The number of dropped actions in each chunk can be calculated as $\lceil f_{a}/f_{p} \rceil$. Similar to the Temporal Ensemble in ACT~\cite{zhao_learning_2023}, overlapping action chunks can lead to multiple predicted actions at the same time step. Following the previous example, at time $t_{N+1}$, we will have both the predicted action $\hat{a}_{N+1}^{t_0}$ at time $t_0$ and $\hat{a}_{N+1}^{t_1}$ at $t_1$. Then we can average these actions to reduce the chance of jitter and increase the smoothness. TEDA does not affect the implementation of the original model and its training procedure but only adapts the inference process, which can be applied to any policy that uses action chunking, such as ACT, Diffusion Policy~\cite{Diffusion_Policy}, etc.


\begin{algorithm}[t]
\small
\caption{Inference with TEDA}
\label{alg:teda_inference}
\begin{algorithmic}[1]
    \STATE Given: trained policy $\pi_\theta$, chunk size $k$, episode length $T_{a}$, weight $w$, prediction frequency $f_{p}$, execution frequency $f_{a}$.
    \STATE Initialize the number of dropped actions $D = \lceil f_{a} / f_{p} \rceil$.
    \STATE Initialize maximum prediction steps $T_{p} = 2 + \lfloor(T_{a} - 1)/D\rfloor$.
    \STATE Initialize a container $B[0 : T_{p}, 0 : C]$, where $B[t_{p}]$ stores actions predicted at step $t_{p}$, and $C = 1 + (T_{p} - 2) \times D + k$.
    \STATE Initialize actions $\hat{a}_{0:k}^{0}$ predicted at $t_{p} = 0$.
    \FOR{timestep $t = 1, 2, \dots, T_{a}$}
        \IF {$t \% D == 1$ }
            \IF {$t == 1$}
                \STATE Add $\hat{a}_{0:k}^{0}$ to $B[0, 0 : k]$ respectively.
            \ELSE 
                \STATE Add $\hat{a}_{t-N:t-N+k}^{t_{p}}$ to $B[t_{p}, t-1:t-1+k]$ respectively.
            \ENDIF
            \STATE Start predicting $\hat{a}_{t:t+k}^{t_{p}}$ with $\pi_\theta$ ($\hat{a}_{t:t+k}^{t_{p}} \mid o_{t}$).
            \STATE $t_{p} = t_{p} + 1$.
        \ENDIF
        \STATE Obtain current step actions $A_{t} = B[t_{x}:t_{y}, t]$, where $t_{x}$ and $t_{y}$ are the indices of the first and last non-zero actions at step $t$ in $B$, respectively.
        \STATE Apply $a_t = \sum_{i} w_i A_t[i]/\sum_{i} w_i$.
    \ENDFOR
\end{algorithmic}
\end{algorithm}


\begin{figure*}[ht]
\centering
\includegraphics[width=0.8\textwidth,trim={0cm 0cm 0cm 1.0cm},clip]{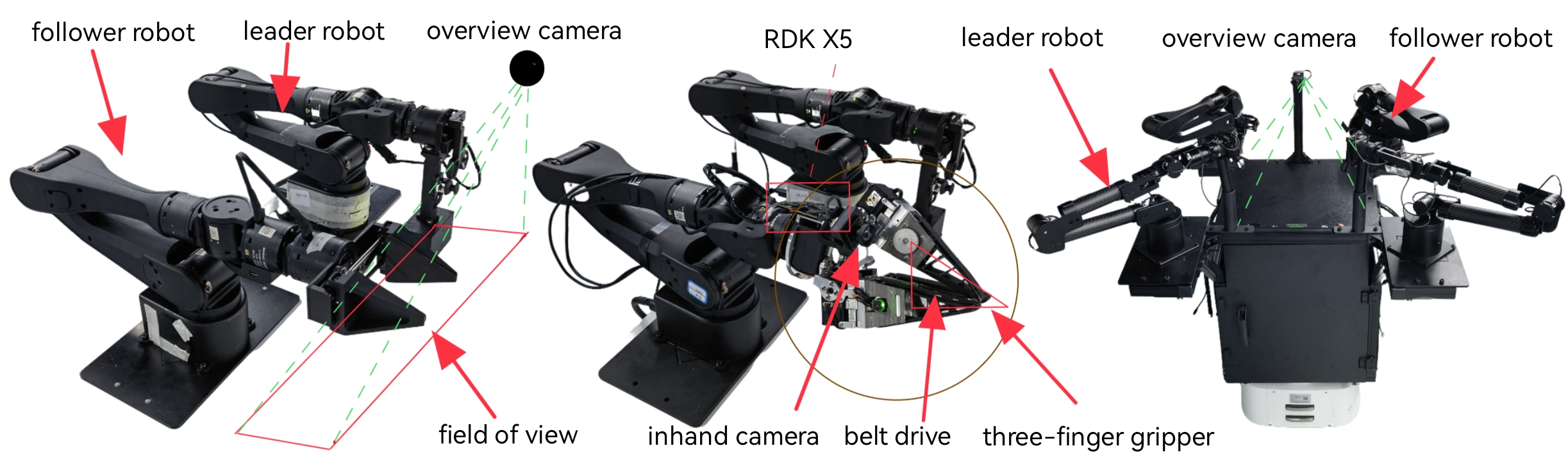}
\caption{\small Overview of teleoperation systems for data collection. \textit{Left:} Single-arm with a two-finger gripper. \textit{Middle:} Single-arm with a three-finger gripper. \textit{Right:} Dual-arm with grippers.}
\label{fig:setup_close}
\end{figure*}

\begin{table*}[ht]
\centering
\caption{Comparison of the hardware specifications between the server and the embedded device.}
\setlength{\tabcolsep}{9pt} 
\renewcommand{\arraystretch}{1.2} 
\begin{tabular}{lcccc}
\toprule
\textbf{Platform} & \textbf{CPU} & \textbf{GPU/BPU} & \textbf{Memory} \\
\midrule
PC & Intel Core i9, 8-core/16-thread, @5.5GHz & NVIDIA RTX 4060, 8GB GDDR6 & 16GB DDR5, 5600MHz \\
\addlinespace
X5 & 8-core Arm Cortex-A55, @1.5GHz & Bayers-architecture BPU @1.0GHz, 0.5GB, 10 TOPS & 2GB/4GB LPDDR4 \\
\bottomrule
\end{tabular}

\label{tab:hardware_comparison}
\end{table*}

\section{Experiment}
In this section, we designed a comprehensive set of experiments to evaluate the performance of the proposed architecture in different manipulation tasks. For each platform and task, we collect data using an operable teleoperation system (as shown in Fig.~\ref{fig:setup_close}) and then train the policy model on the server (PC) with the collected measurements. After that, the policy is transferred to an embedded platform (RDK X5)\footnote[1]{https://developer.d-robotics.cc/rdkx5} with the proposed pipeline. Details regarding the system specifications and resources can be found in Tab.~\ref{tab:hardware_comparison}. 

\subsection{Manipulation Platforms and Tasks}
We designed three distinct hardware systems (as illustrated in Fig.~\ref{fig:setup_close}) for performing various manipulation tasks: a system with a single arm mounted with a two-finger gripper for tasks that require simple and precise grasping actions, a system with dual arms mounted with two-finger grippers for more complex tasks that require coordinated movement between the two arms, and a system with a single arm mounted with a three-finger gripper for tasks that require fine motor control. The robotic arms, both leaders (used in data collection) and followers, utilized in our setup were 6-DoF AIRBOT \footnote[2]{https://airbots.online/}Play robotic arms.

\begin{figure*}[t!]
\centering
\includegraphics[width=0.95\textwidth,trim={0cm 0cm 0cm 0cm},clip]{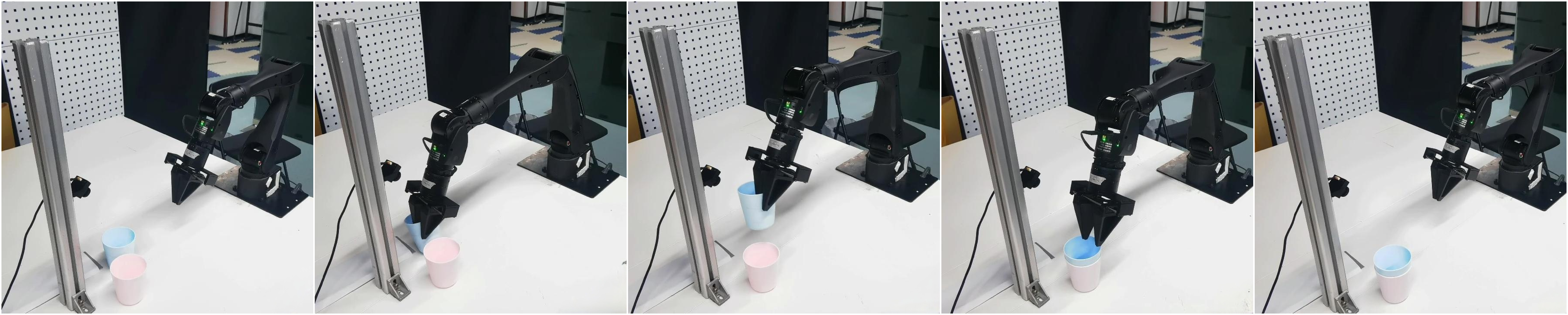}
\caption{
\textit{Task 1: Single-Arm Cup Stacking:} \small This task involves using a single robotic arm with a two-finger gripper to place a blue cup into a pink cup. First, the robotic arm approaches the blue cup and picks it up (Subtask 1: Grasp). Then, the arm adjusts the position of the blue cup to place it into the pink cup. Finally, the arm opens the gripper and returns to the initial position (Subtask 2: Place).}
\label{fig:task1}
\vspace{0.3cm}

\includegraphics[width=0.95\textwidth,trim={0cm 0cm 0cm 0cm},clip]{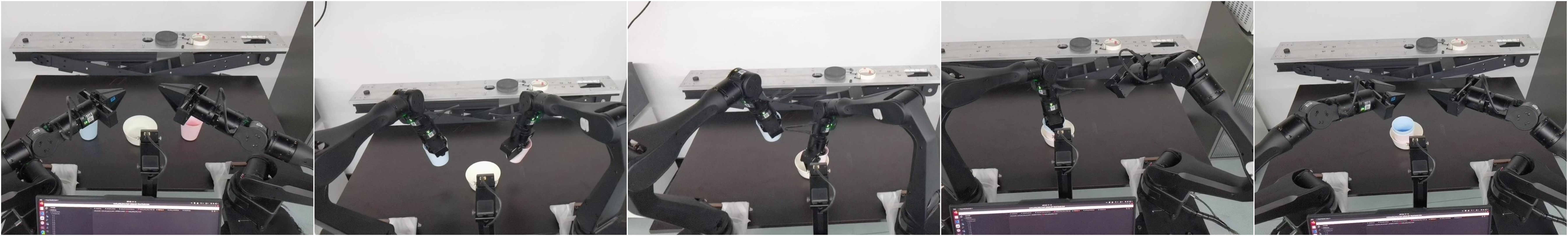}
\caption{
\textit{Task 2: Dual-Arm Cup Stacking:} \small This task involves using a two-arm system to stack cups into a bowl. Both robotic arms simultaneously approach these two cups, with each arm executing specific subtasks. First, the right arm grasps the pink cup (Subtask 1) while the left arm grasps the blue cup (Subtask 3). Once the cups are secured, both arms move the cups over the bowl. The right arm then places the pink cup into the bowl, opens its gripper, and returns to the initial position (Subtask 2). Following this, the left arm places the blue cup inside the pink cup, opens its gripper, and also returns to its initial position (Subtask 4). } 
\label{fig:task2}
\vspace{0.3cm}
\includegraphics[width=0.95\textwidth,trim={0cm 0cm 0cm 0cm},clip]{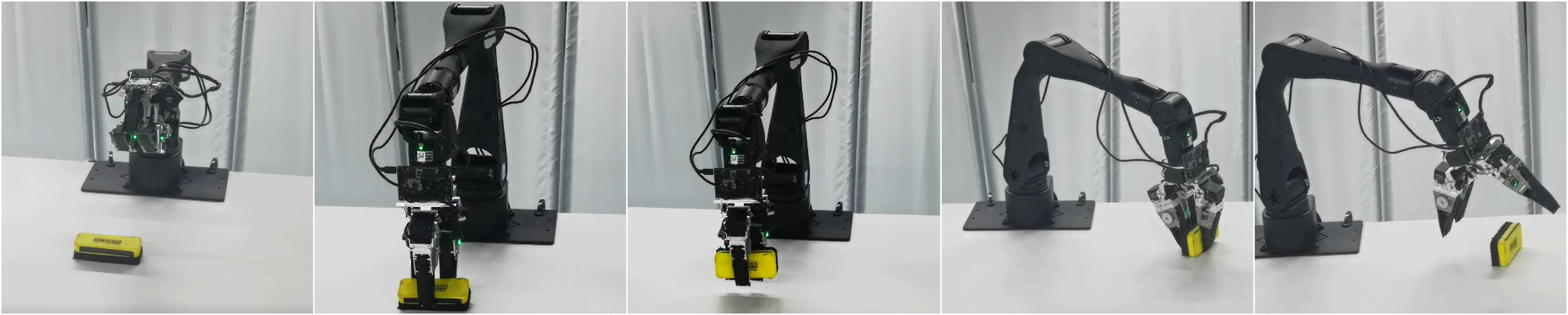}
\caption{
\textit{Task 3: Whiteboard Eraser Rotation:} \small In this task, a single robotic arm mounted with a three-finger gripper is used to manipulate a whiteboard eraser. The arm first picks up the whiteboard eraser (Subtask 1: Grasp the board eraser), then rotates it using the conveyor belt of the three fingers to change its orientation (Subtask 2: Rotate the board eraser). Finally, the arm shifts the eraser to a specific location, activates the conveyor belt to place the eraser down, and then opens the gripper to complete the task (Subtask 3: Place the eraser).}
\label{fig:task3}
\end{figure*}

Tasks for evaluation are illustrated in Fig.~\ref{fig:task1},~\ref{fig:task2}, and~\ref{fig:task3}, where we break down the tasks into more specific subtasks. These tasks encompass a series of operations, including controlling the movement of robotic arms, managing the grasping actions of grippers, and manipulating the rotational motion in some of the cases. 


\subsection{Data Collection and Policy Training}
We utilized a teleoperation system for data collection, where operator-controlled leader arms guided follower arms to gather precise and effective data. The dataset comprises joint positions of both leader and follower robots, along with RGB images (each at $480 \times 640$ resolution). Considering the complexity of the tasks, we recorded 120, 1000, and 400 timesteps for tasks 1 $\sim$ 3, respectively, resulting in 300, 50, and 50 demonstration episodes. The sampling frequency was set at 25 Hz for all tasks. To improve spatial generalization, objects for manipulation were randomly positioned within a 5 cm $\times$ 5 cm area during each data collection process. 

The imitation learning model for evaluation in our experiments is the ACT model~\cite{Zhao2023LearningFB}, which includes both the transformer and CNN structure. All models are trained with the same hyperparameters and the chunk size is 25. 

\subsection{Experiment Results}
    To evaluate the importance of different factors of our proposed structure, we first did the ablation tests on the Input Shape Unification (ISU) component. It is important to note that Symmetric Quantization (SQ) is essential, as the model cannot run on the X5 without it due to hardware limitations. Thus we are showing the effect of ISU in the ablation test. Then we did the overall evaluation among all the scenarios to complete various manipulation tasks. 

\subsubsection{With/Without ISU}
In this study, we evaluate the impact of ISU on performance across three settings: baseline inference on the PC, SQ on the X5, and SQ combined with ISU on the X5. The accuracy and prediction time of the compressed models are presented in Table~\ref{tab:cp_model_info}. The results indicate that the model compressed solely with SQ requires approximately 47 times longer to perform inference on the X5 compared to the uncompressed model running on a PC. However, by incorporating ISU, the inference time on the X5 is reduced by a factor of 6, without any loss in model accuracy. This demonstrates the effectiveness of ISU in optimizing inference time while maintaining model performance. Consequently, the model with SQ and ISU will be used for all subsequent experiments on the X5 platform.

\begin{table}[ht]
\centering
\caption{Comparison results for with/without ISU. The prediction accuracy of the model is calculated as $e_{0}/e_{1}$, where $e_{0}$ and $e_{1}$ are the MSE values between the predicted results and the ground truth for the model before and after compression, respectively among 25 trails of data.}
\setlength{\tabcolsep}{4.5pt} 
\renewcommand{\arraystretch}{1.2} 
\begin{tabular}{lcccc}
\toprule
\textbf{Platform} &\textbf{Method} & \textbf{Accuracy} & \textbf{Prediction time} \\
\midrule
PC & --- & 1 & 0.013 \\
\addlinespace
X5&SQ & 0.997 & 0.614 \\
\addlinespace
X5&SQ + ISU & 0.997 & 0.103\\
\bottomrule
\end{tabular}

\label{tab:cp_model_info}
\end{table}

\begin{table}[htbp]
    \centering
    \setlength{\tabcolsep}{4.5pt}
    \renewcommand{\arraystretch}{1.3}
    \caption{\small Comparison of inference frequency. $t_{1}$: observation time, $t_{2}$: prediction time, $t_{3}$: communication time, $t_{4}$: execution time. For the baseline, the total inference time is the sum of all four components ($t_{1}$ $t_{2}$ $t_{3}$ $t_{4}$). In contrast, for ours (policy with TEDA), the total inference time is only the sum of $t_{3}$ and $t_{4}$ since the prediction and execution are parallel.}
    \begin{tabular}{cccccc}
        \toprule
        \multirow{2}{*}{\textbf{Platform}} & \multicolumn{4}{c}{\textbf{Time per step (s)}} & \multirow{2}{*}{\textbf{Total Inference Time}} \\
        \cline{2-5}
         & $t_{1}$ & $t_{2}$ & $t_{3}$ & $t_{4}$ &  \\
        \midrule
        \textbf{PC (Baseline)} & 0.013 & 0.012 & 0.001 & 0.040 & 0.066 \\
        \textbf{X5 (Baseline)} & 0.017 & 0.103 & 0.001 & 0.040 & 0.161 \\
        \textbf{X5 (Ours)} & 0.017 & 0.103 & 0.001 & 0.040 & 0.041 \\
        \bottomrule
    \end{tabular}
    \label{table:cost_time}
\end{table}

\subsubsection{Overall Task Results}
As introduced above, we designed three tasks to evaluate our pipeline. For each task, we compare the results across three scenarios: inference on a PC (\textbf{PC (Baseline)}), direct inference on the embedded device (\textbf{X5 (Baseline)}), and deployment on the embedded device using our proposed pipeline (\textbf{X5 (Ours)}). Each test was repeated 25 times to avoid bias. 

As shown in Table~\ref{table:success_rate} and Table~\ref{table:cost_time}, the policies evaluated on the PC consistently achieved the highest success rates across all tasks and subtasks. In contrast, the success rate of the baseline strategy on the X5 embedded device is significantly lower. A primary cause of task failure, as observed, is the jitteriness in motion, attributed to the low inference frequency during the robotic arm's movement. Additionally, the low inference frequency hinders the system's ability to make timely adjustments to changes in the scene, which further reduces the success rate. After incorporating the proposed pipeline, we can achieve a similar inference performance as PC does in terms of the success rate and completion time, especially for the Dual-Arm Cup Stacking task and the Whiteboard Eraser Rotation task. The results showcase the efficiency of our approach in more complex, multi-step tasks.
\begin{table*}[t!]
\centering
\setlength{\tabcolsep}{10pt} 
\renewcommand{\arraystretch}{1.5} 
\caption{\small Success rate (\%) comparison for the tasks with three scenarios. }
\begin{tabular}{cccccccccc}
\toprule
\textbf{\multirow{2}{*}{\textbf{Platform}}} & \multicolumn{2}{c}{\textbf{Task 1}} & \multicolumn{4}{c}{\textbf{Task 2}} & \multicolumn{3}{c}{\textbf{Task 3}} \\ 
\cmidrule(lr){2-3}
\cmidrule(lr){4-7}
\cmidrule(lr){8-10}

 & Grasp & Place & Subtask1 & Subtask2 & Subtask3 & Subtask4 & Subtask1 & Subtask2 & Subtask3 \\ 
\midrule
\textbf{PC (Baseline)}  &92 &84 &88 &80 &84 &72 &76 &72 &56 \\ 
\addlinespace
\textbf{X5 (Baseline)}  &68 &56 &64 &52 &60 &48 &60 &56 &44 \\ 
\addlinespace
\textbf{X5 (Ours)}  &88 &76 &80 &72 &80 &64 &72 &68 &52 \\ 
\bottomrule
\end{tabular}

\label{table:success_rate}
\end{table*}

\section{Limitations and Discussions}
This paper presents a method for deploying high-performance algorithms on resource-constrained edge devices. By employing Input Shape Unification (ISU), we significantly improved the inference frequency of quantized models on the X5 embedded platform. Additionally, we introduced a practical asynchronous parallel method, Temporal Ensemble with Dropped Actions (TEDA), which ensures continuity during action execution. Experimental results validate the effectiveness of our approach, with success rates approaching those observed on PC platforms.

However, when the model prediction frequency is much lower than the action execution frequency, the number of dropped actions in TEDA will be large, which requires the model to have a very large chunk size. However, the maximum value of chunk size is limited by the episode length of the specific task, and a large chunk size may also potentially reduce the success rate of the task~\cite{Zhao2023LearningFB}, so TEDA may no longer be applicable in this case.

\bibliographystyle{IEEEtran}
\bibliography{airbot}

\end{document}